\documentclass[lettersize,journal]{IEEEtran}
\usepackage{amsmath,amsfonts}
\usepackage{algorithm}
\usepackage{array}
\usepackage[caption=false,font=normalsize,labelfont=sf,textfont=sf]{subfig}
\usepackage{textcomp}
\usepackage{stfloats}
\usepackage{url}
\usepackage{verbatim}
\usepackage{graphicx}
\usepackage{cite}
\usepackage{multirow}
\usepackage[numbers]{natbib}
\usepackage{amssymb}
\usepackage{algpseudocode}
\usepackage{algorithm}
\usepackage{float}

\usepackage[table]{xcolor} 
\usepackage{booktabs}       
\usepackage{soul}           
\usepackage{graphicx}   
\usepackage{array, booktabs}

\begin{document}

\title{PRIDE: Privileged Information-enhanced Distillation for Empathetic Dialogue Generation}

\author{Jiaqiang Wu, Zhouan Zhu, *Shangfei Wang, 
\thanks{This work has been supported by projects from the National Natural Science Foundation of China (62376255). Shangfei Wang is the corresponding author. Jiaqiang Wu, Shangfei Wang, Yanan Chang, and Zhouan Zhu are with Anhui Robot Technology Standard Innovation Base, School of Computer Science and Technology, University of Science and Technology of China, Hefei, Anhui 230027, PR China. E-mail: jqwu@mail.ustc.edu.cn; sfwang@ustc.edu.cn; cyn123@mail.ustc.edu.cn; zza2021@mail.ustc.edu.cn}
}

\markboth{Journal of \LaTeX\ Class Files,~Vol.~14, No.~8, August~2021}%
{Shell \MakeLowercase{\textit{et al.}}: A Sample Article Using IEEEtran.cls for IEEE Journals}


\maketitle

\begin{abstract}
Large language models have demonstrated significant capabilities in generating diverse and context-aware responses for empathetic dialogue. However, their computational demands severely limit their deployment in resource-constrained environments. 
While knowledge distillation offers a promising compression solution, it often fails to transfer the nuanced understanding essential for empathy, as it overlooks the implicit contextual cues that guide human connection. To bridge this gap, we propose a \textbf{pr}ivileged \textbf{i}nformation-enhanced knowledge \textbf{d}istillation method for \textbf{e}mpathetic dialogue generation (PRIDE). 
Our method leverages privileged information, such as expert psychological annotations or future event summaries, which is available exclusively during training but unavailable at inference time. This allows us to transfer the teacher model's empathetic reasoning to smaller models without relying on extra inputs during deployment.
Specifically, PRIDE has three key components: (1) An empathy-reasoning prompt that guides the teacher to explicitly decompose the empathetic process into understanding feelings and analyzing situations step-by-step; (2) A multi-source attention mechanism that directs the student to effectively integrate privileged information; (3) A dual-alignment loss that combines reversed Kullback-Leibler divergence and maximum mean discrepancy to ensure robust knowledge transfer at both logit and feature levels. Experiments on multi-modal and text-only datasets demonstrate that our method achieves competitive performance, and in some cases matches or even surpasses larger teacher models in terms of accuracy and semantic relevance. Code will be released once the paper is accepted.
\end{abstract}


\section{Introduction}
Empathy is the capacity to comprehend and share the feelings of others  \citep{macarov}, which is a cornerstone of human communication. It is often deconstructed into two key components: affective empathy (feeling with the user) and cognitive empathy (understanding the user's situation) \cite{damon2006handbook}. Research in empathetic dialogue generation (EDG) domain has largely followed these two aspects, with early works focusing on affective mimicry \cite{mime,moel,empdg} and later works enhancing cognitive understanding through external knowledge \citep{cem,seek,case} or modeling self-other distinctions \citep{empsoa}.
\begin{figure}[t]
	\centering
	\includegraphics[width=0.95\linewidth]{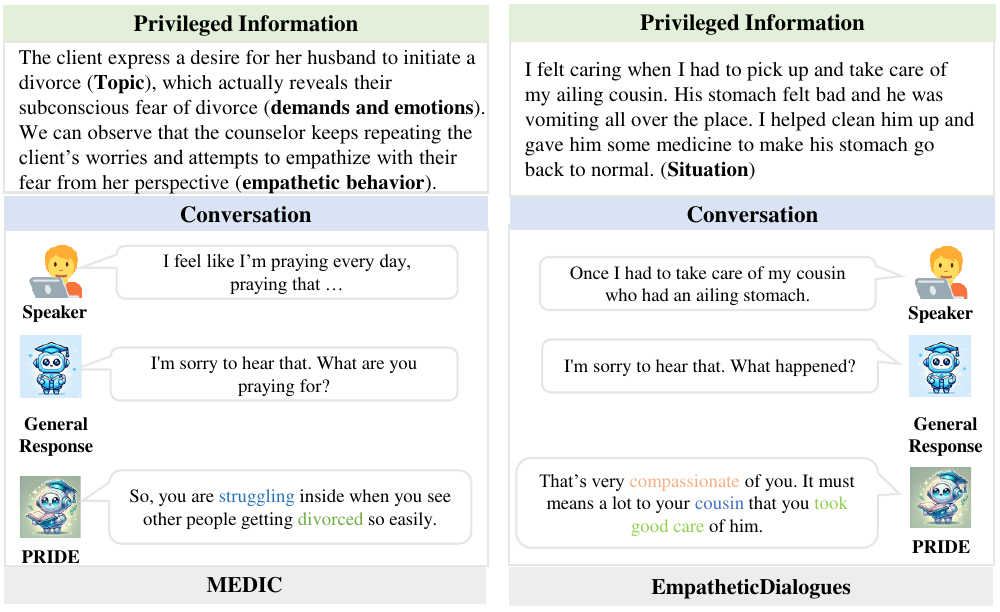}
	\caption{Examples of privileged information.}
	\label{example_0}
\end{figure}
The recent advent of large language models (LLMs) has marked a significant leap forward, enabling the generation of empathetic responses with superior diversity, fluency, and contextual awareness \citep{yan-etal-2024-talk,stickerconv,empathyear,zhao2023chatgptequip}. However, their immense computational cost presents a formidable barrier to practical deployment in resource-constrained environments. To address this, knowledge distillation (KD) emerges as a compelling strategy, which is designed to transfer the capabilities of large teacher models to smaller and more efficient student models \cite{xu2024surveykd,hinton2015distilling}.

Despite its promise, standard KD often fails to transfer the nuanced understanding essential for genuine empathy. Conventional methods typically focus on mimicking final outputs, overlooking the latent reasoning processes and contextual signals crucial for empathetic dialogue. The essence of empathy lies in grasping a speaker's underlying emotional state, which is often unstated in the dialogue itself. This ambiguity frequently leads previous models to produce generic and superficial responses. To bridge this gap, we introduce a privileged information-enhanced distillation method for empathetic dialogue generation (PRIDE).
Our method is rigorously grounded in the Learning Using Privileged Information (LUPI) framework \cite{vapnik2015learning}. LUPI formalizes a teaching paradigm in which a student learns from auxiliary data (privileged information) that is available only during the training phase. 
We argue that in real-world empathetic interactions, certain critical insights, such as a counselor's post-session psychological analysis or a retrospective summary of an event, are often inaccessible or delayed in real-time scenarios. However, this information is extremely valuable for learning. Including such future or expert knowledge at test time would constitute data leakage or be practically infeasible. While ignoring it wastes a valuable supervision signal. Therefore, we investigate a new direction: Can we use privileged information as a training-only catalyst to improve the student's empathetic understanding?
While LUPI has been influential, its application to empathetic dialogue remains unexplored. Instead of focusing on explicit knowledge injection or architectural engineering, which are common in knowledge-enhanced empathetic dialogue \cite{survey2023}, we investigate a new direction: Can we use privileged information as a catalyst during distillation to improve the student's empathetic understanding? In this work, we utilize two types of privileged information: (1) expert-provided high-level analyses of clients' mental states (from the MEDIC dataset) and (2) predefined situation descriptions that outline the real-life context triggering a conversation (from the EmpatheticDialogues dataset).
We leverage this information not as an input for the model during inference, but as a teacher's ``explanation'' to guide knowledge transfer during training.
Therefore, we integrate this privileged information to help student models better capture the emotional understanding and situational reasoning required for empathetic dialogue generation. Figure \ref{example_0} illustrates how the privileged information provides emotional and situational cues beyond dialogue text. 

In contrast to prior works that build separate modules \cite{empdg,kemp,empsoa}, we demonstrate that by distilling knowledge from a teacher model guided by privileged information, a student model can develop a more robust empathetic capability.
Our method begins by employing an empathy-reasoning prompt to compel the teacher model to explicitly reason about the speaker's emotional state and situation before generating a response.
This ensures the teacher's output is grounded in a logical empathetic pathway.
We then equip the student model with a multi-source attention mechanism, allowing it to integrate information from the dialogue context and the privileged information during training.
Finally, to guarantee a faithful transfer of knowledge, we enforce alignment at the feature and logits levels using maximum mean discrepancy (MMD) \cite{mmd} and reversed Kullback-Leibler divergence (RKLD) \cite{gu2024minillm}, respectively.

Our primary contributions are:
\begin{itemize}
	\item {To the best of our knowledge, we are the first to leverage privileged information for knowledge distillation in empathetic dialogue.}
	\item {We propose a set of synergistic techniques, including an empathy-reasoning prompt for the teacher, a multi-source attention mechanism for the student, and a dual-level alignment to ensure robust knowledge transfer.}
	\item {Experiments on multi-modal and text-only datasets validate our method's scalability and superiority. Notably, we demonstrate that a smaller student model can achieve performance comparable to or better than its teacher by effectively internalizing privileged insights.}
\end{itemize}

\begin{figure*}[htbp]
	\centering
	\includegraphics[width=0.95\linewidth]{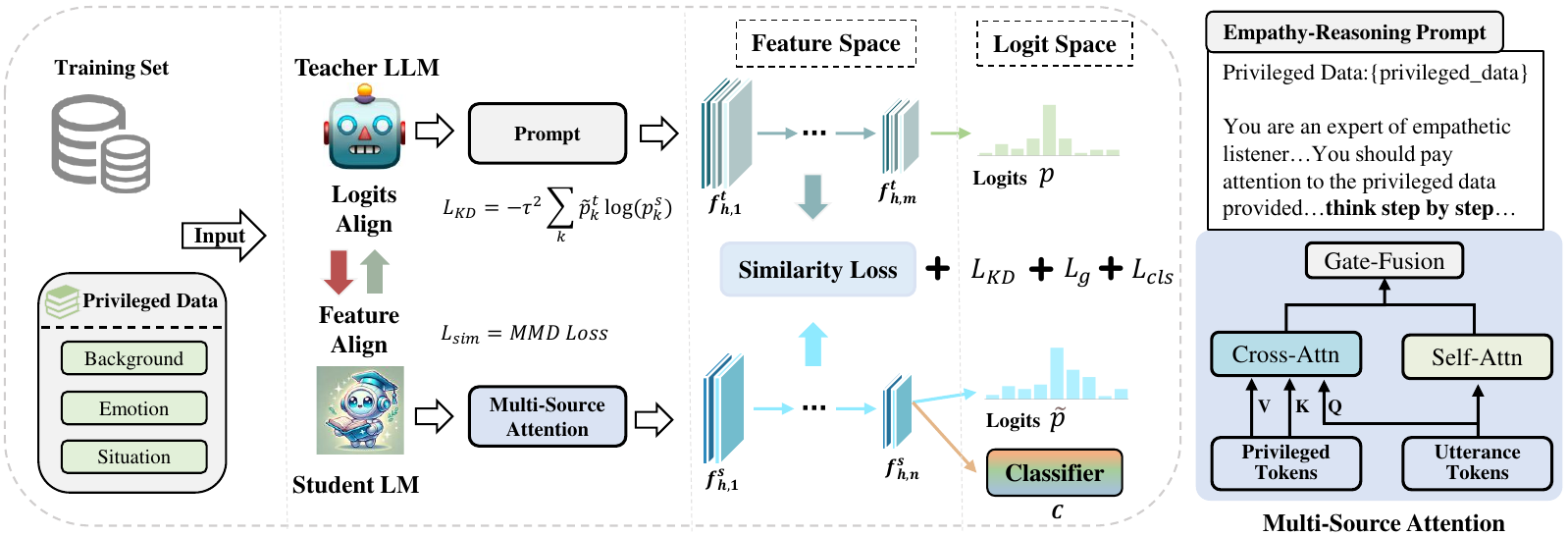}
	\caption{Overview of our method. During training phase, privileged information and training data are provided as inputs, whereas privileged information is excluded during the inference phase. Specifically, $L_{KD}$, $L_{sim}$, $L_{cls}$, and $L_g$ represent the loss functions for logits alignment, feature alignment, emotion recognition, and response generation, respectively. $f_{h, m}^t$ and $f_{h, n}^s$ are the intermediate representations from the teacher's $m$-th and the student's $n$-th hidden layer, respectively.}
	\label{framework}
\end{figure*}

\section{Related Work}

\subsection{Knowledge Distillation of LLMs}

Knowledge distillation is a paradigm for training a compact student model under the guidance of a larger teacher model \cite{hinton2015distilling,rusu2015policy,gou2021knowledge}. In the era of LLMs, KD has become a critical technique for transferring the comprehensive capabilities of LLMs to smaller models. Data augmentation has emerged as a popular technique, where a teacher LLM is prompted with a small set of examples to generate a larger, task-specific dataset for training the student. Researchers have also explored self-distillation methods \cite{compress,selfreward}, where open-source LLMs act as their own teachers to refine and enhance their capabilities. By employing these varied distillation techniques, the performance gap between proprietary and open-source models is narrowed, making the advanced abilities of LLMs more accessible \cite{xu2024surveykd}.

While these distillation techniques are effective for transferring general language abilities, they fall short in capturing the emotional and contextual nuances required for empathy. To address this limitation, we enhance the distillation process with privileged information, which provides a richer and more context-aware learning signal that enables the student model to better inherit the teacher's empathetic abilities.

\subsection{Empathetic Dialogue Generation}
Empathy is a multi-dimensional construct comprising both affective (sharing feelings) and cognitive (understanding situations) aspects \citep{aff-cog}. To capture affective empathy, models have been equipped with emotion recognition capabilities, such as using mixtures of specialized encoders \citep{moel}, classifying emotion polarity \citep{mime}, or identifying feelings at utterance and token levels \citep{empdg}. Concurrently, to enhance cognitive empathy, numerous works enrich models' situational understanding. A prominent approach uses knowledge graphs like COMET \cite{comet} to infer context from dialogue history \citep{cem, kemp, case}, while subsequent efforts have addressed the resulting challenges, like discrepancies between inferred knowledge and perceived emotions \cite{seek}. Further architectural innovations include mechanisms for self-other differentiation \citep{empsoa} and iterative attention to better capture emotionally salient words \citep{yang-iterative}.

Prior work has modeled empathy using static knowledge graphs \cite{comet,speer2017conceptnet} or dedicated modules, but these approaches are inherently limited by the information explicitly contained in dialogue text. However, empathetic dialogue often relies on implicit emotional states and situational contexts that are not directly expressed. To address this gap, we introduce privileged information, including predefined situation descriptions in EmpatheticDialogues \cite{empatheticdialogues} and expert-provided supervisory feedback in MEDIC \cite{medic}, which function as a teacher's explanation during training. These signals provide explicit guidance that helps the student model acquire emotional understanding and situational reasoning beyond surface-level dialogue content.

\section{Problem Statement}
Our task is to generate an empathetic response $Y = \{y_1, \ldots, y_m\}$ by transferring knowledge from a teacher model $f^t$ to a student model $f^s$. This task is performed given a dialogue context $\mathcal{D} = \{U_1, \ldots, U_{n}\}$ and corresponding privileged information $\mathcal{P}$, where $n$ is the number of utterances, and each utterance $U_i = \{w_1^{i}, \ldots, w_{n_i}^{i}\}$ is a sequence of $n_i$ words. For multi-modal datasets, the input is augmented with visual data $\mathcal{V} = \{v_1, \ldots, v_n\}$, where each image $v_j$ is paired with the utterance $U_j$.

\section{Methodology}
As Figure \ref{framework} shows, our method distills a teacher LLM into a student model with the help of privileged information. The teacher is guided by a empathy-reasoning prompt to reason about the speaker's emotions and situation. The student employs a multi-source attention module to fuse dialogue and privileged information, and we enforce dual-level alignment: matching their features via maximum mean discrepancy loss and aligning their logits with reverse KL divergence to focus on the teacher's confident predictions. An auxiliary classifier for emotion further grounds the learning.

\subsection{Empathy Reasoning Prompt}
To leverage the teacher LLM's understanding abilities, we design an empathy-reasoning prompt \cite{cot}. Our design is motivated by the nature of empathy: one must first comprehend the speaker's feelings (affective empathy) and their situation (cognitive empathy) before an empathetic response can be generated. The prompt guides the teacher model to first output its analysis of the speaker's emotion and situation, and then generate the empathetic response. This serves as a transparent learning target for the student model. 
The complete prompt is provided in Appendix.

\subsection{Learning From privileged information}
Our method facilitates knowledge transfer by enabling the student model to learn directly from privileged information. To achieve this, we employ a multi-source attention module with a gated fusion mechanism. 
Crucially, we treat the dialogue context $\mathcal{D}$ and privileged information $\mathcal{P}$ as two separate input streams rather than concatenating them. This separation is vital to prevent the student model from simply memorizing and overfitting to the privileged information, which is unavailable during inference. 

First, we compute the self-attention output $A_\mathcal{D}$ for the dialogue context. We also compute the cross-attention output $A_\mathcal{P}$, where the query comes from the dialogue context while the keys and values come from the privileged information:
\begin{equation}
	\begin{aligned}
		A_\mathcal{D} &= \sigma \left(\frac{Q_\mathcal{D}K_\mathcal{D}^T}{\sqrt{d_k}}\right)V_\mathcal{D} \
		\\
		A_\mathcal{P} &= \sigma \left(\frac{Q_\mathcal{D}K_\mathcal{P}^T}{\sqrt{d_k}}\right)V_\mathcal{P}
	\end{aligned}
	\label{attention}
\end{equation}
where $Q_\mathcal{D}, K_\mathcal{D}, V_\mathcal{D} \in \mathbb{R}^{l \times d_k}$ are query, key, value matrices for the dialogue context, $\sigma$ is the softmax function, $K_\mathcal{P}, V_\mathcal{P} \in \mathbb{R}^{l_p \times d_k}$ are the key and value matrices for the privileged information. $A_\mathcal{D} \in \mathbb{R}^{l \times d_k}$ represents what the model understands from the dialogue alone, and $A_\mathcal{P} \in \mathbb{R}^{l \times d_k}$ represents the contribution of privileged information. $d$ is the dimension of the embedding layers, $d_k$ is the dimension of the attention layers, $l$ and $l_p$ are the length of the dialogue sequence and the privileged information, respectively.

Then, a gating mechanism is employed to dynamically fuse the outputs from the self-attention and the cross-source attention modules:
\begin{equation}
	A_s = \delta (W_g A_\mathcal{D}) \odot A_\mathcal{D} \; + \; (1 - \delta(W_g A_\mathcal{D})) \odot A_\mathcal{P}
\end{equation}
where $A_s \in \mathbb{R}^{l \times d_k}$ is the result of the fused multi-source attention, $W_g \in \mathbb{R}^{d_k \times d_k}$ are the weights of the gating mechanism, $\delta (\cdot)$ represents the sigmoid function, and $\odot$ denotes element-wise multiplication.

During inference, the input representations for privileged information are initialized as zero vectors (i.e., null states). 
Mathematically, this implies that the Key and Value matrices for the privileged stream become zero matrices ($K_\mathcal{P} = \mathbf{0}, V_\mathcal{P} = \mathbf{0}$). 
Consequently, the cross-attention term $A_\mathcal{P}$ vanishes:
\begin{equation}
	A_\mathcal{P} = \sigma \left(\frac{Q_\mathcal{D} \cdot \mathbf{0}^T}{\sqrt{d_k}}\right) \cdot \mathbf{0} = \mathbf{0}
\end{equation}
Substituting this into Eq. (2), the fusion mechanism automatically simplifies to:
\begin{equation}
	A_s = \delta (W_g A_\mathcal{D}) \odot A_\mathcal{D}
\end{equation}
In this scenario, the multi-source attention module effectively degenerates into a gated self-attention mechanism. 
The gate $\delta (W_g A_\mathcal{D})$ dynamically scales the dialogue context based on the model's confidence, ensuring that the student model can generate coherent responses solely based on the dialogue context $\mathcal{D}$ without requiring architectural changes or auxiliary inputs. 
To bridge the discrepancy between the training phase and the inference phase, we introduce a privileged dropout strategy. Specifically, during training, we randomly mask the privileged information by setting its key and value matrices to zero with a probability $p_{drop}$. This strategy prevents the student model from becoming overly reliant on the privileged branch, and it forces the model to actively mine and verify cues in the dialogue context to reconstruct the missing privileged insights. Thus, the gating mechanism $\delta$ learns robust feature fusion weights that can dynamically adapt to the absence of privileged signals during inference.

The multi-source attention mechanism enriches the dialogue representation with contextual insights from the privileged information, thereby enabling the model to generate more relevant and empathetic responses. The fused attention output $A_s$ is then propagated through the subsequent decoder layers to produce the final hidden states $h'_t$, which are used for token generation.
The probability distribution over the vocabulary for the $t$-th token is computed as:
\begin{equation}
	p(y_t | y_{< t}) = \sigma \,(W_h h'_t + b)
\end{equation}
where $W_h \in \mathbb{R}^{\lvert V \rvert \times d_k}$ is the vocabulary projection matrix, $V$ is the vocabulary size, and $b$ is bias. We adopt the standard negative log-likelihood (NLL) loss for training:
\begin{equation}
	\mathcal{L}_g = \mathbb{E}_{(\mathcal{D},\mathcal{P},Y)} [- \sum^N_{t=1} p(y_t | y_{<t})]
\end{equation}
Furthermore, to ensure the generated responses are emotionally appropriate, an auxiliary classifier is employed. It leverages the hidden states $h'$ to predict emotion labels. We use the cross-entropy loss to train the classifier:
\begin{equation}
	\begin{aligned}
		\hat{e} &= \sigma (W_e \cdot LN\,(h'_t)) \\
		\mathcal{L}_{cls} &= - \sum_{i=1}^{C} e_i \log {\hat{e}_i}
	\end{aligned}	
\end{equation}
where $LN$ denotes layer normalization, $W_e \in \mathbb{R}^{C \times d_k}$ are trainable weights of the classifier, $C$ is the number of classes, $\hat{e}$ is the predicted probability distribution and $e$ is the one-hot encoded ground-truth label.

\subsection{Alignment in Feature Space}
Different model layers are known to capture features at varying levels of abstraction, from local syntax to global semantics. We therefore apply distillation at multiple levels of the model's architecture. We enforce this multi-level alignment by minimizing a similarity loss between the intermediate and final hidden states of the teacher and student models. For this purpose, we employ the MMD loss \cite{mmd}, because it is well-suited for empathetic dialogue as it measures the divergence between probability distributions in a high-dimensional space, allowing it to align complex and non-linear relationships within the teacher and student models' features. It enables the student model to mimic the teacher's nuanced semantic understanding across different hierarchical levels of representation. Given the teacher's hidden states $f_{h, i}^t$ from layer $i$ and the student's hidden states $f_{h, j}^s$ from layer $j$, the similarity loss is formulated as:
\begin{equation}
	\begin{aligned}
		L_{sim} &= \mathbb{E}[k(\Phi_t(f_{h, i}^t), \Phi_t(f_{h, i}^t))] \\ 
		&+ \mathbb{E}[k(\Phi_s(f_{h, j}^s), \Phi_s(f_{h, j}^s))] \\
		&- 2 \cdot \mathbb{E}[k(\Phi_t(f_{h, i}^t), \Phi_s(f_{h, j}^s))] \\
	\end{aligned}
	\label{sim}
\end{equation}
where $(i,j) \in \{(\lfloor \frac{m}{2} \rfloor, \lfloor \frac{n}{2} \rfloor), (m, n)\}$, $m$ and $n$ are the number of hidden layers in the teacher and the student models, respectively. $\Phi_t(\cdot)$ and $\Phi_s(\cdot)$ are linear transformation functions, which are applied when the hidden states of the teacher and student models have different dimensions. The function $k$ represents the Gaussian RBF kernel.

\subsection{Alignment in Response}
In addition to aligning the models in feature space, we also distill knowledge at the output level. Specifically, we align the student's output distribution with the teacher's by minimizing the RKLD between their predicted logits. 
Unlike the mean-seeking nature of forward KLD, which can average a teacher's predictions into general responses, RKLD is mode-seeking \cite{wang2025abkd}. It compels the student to focus on the teacher's most confident predictions. We hypothesize that effective empathy requires a clear stance rather than a vague one. RKLD's mode-seeking property forces the student to learn the teacher's most unambiguous empathetic expressions, thus preventing general responses.

Concretely, given the input dialogue context $\mathcal{D}$, the teacher model produces a logit vector $p = [p_1, p_2, \ldots, p_K]$ that is then softened using a temperature scaling hyperparameter $\tau$ to create a soft target probability distribution:
\begin{equation}
	p^t_k = \frac{e^{(p_i / \tau)}}{\sum_j e^{p_j / \tau}}
\end{equation}
Similarly, the student model produces its own probability distributions $\tilde{p}^s_k$. The knowledge distillation loss is then defined using RKLD:
\begin{equation}
	L_{KD} = \tau^2 \sum_{k=1}^{|V|} \tilde{p}^s_k \left( \log \tilde{p}^s_k - \log \tilde{p}^t_k \right)
	\label{kl}
\end{equation}
Optimizing Equation \ref{kl} thus aligns the student model's logits with those of the teacher model.

\subsection{Overall Loss}
We integrate guidance from feature space and the generated response, incorporating both classification loss and generation loss. The overall loss is defined as follows:
\begin{equation}
	\mathcal{L} = \alpha\mathcal{L}_{sim} + \beta\mathcal{L}_{KD} + \gamma\mathcal{L}_{cls} + \delta\mathcal{L}_{g}
\end{equation}
where $\alpha, \beta, \gamma, \delta$ are hyper-parameters.

\section{Experiments}
\subsection{Datasets}

\textbf{MEDIC} is a multimodal empathetic dataset designed to advance computational empathy understanding in psychological counseling sessions. The dataset comprises 771 video clips collected from the counseling sessions. Its privileged information is derived from the evaluative feedback provided by annotators in the dataset.

\textbf{EmpatheticDialogues} (ED) is a text-based empathetic dialogue corpus comprising 24,850 one-to-one open-domain conversations. The privileged information consists of the situational information provided with each dialogue, which describes the event that prompted the conversation.

\subsection{Model Families} 
We conducted experiments across three open-source model families: Qwen2.5-VL \citep{Qwen2.5-VL}, LLaVA \cite{hinck2024llavagemma}, and Gemma3 \cite{gemma_2025}. We chose Qwen2.5-VL (7B), LLaVA (7B), and Gemma3(12B) as teacher models. For student models, we used Qwen2.5-VL (3B), LLaVA (2B), and Gemma3 (4B, 1B). Each teacher model was fine-tuned on our experimental datasets in advance. Specifically, when experimenting on the text-only EmpatheticDialogues dataset, to maintain architectural consistency across multi-modal and text-only experiments, we exclusively utilize the language model backbone of these architectures (bypassing the visual encoders) to ensure the setup is appropriate for the data modality.

\subsection{Baselines}
\begin{table*}[]
	\centering
	\scriptsize
	\renewcommand{\arraystretch}{1.0}
	\caption{Automatic evaluation results of our method. F\textsubscript{BERT}: F1 score for BERTScore, Emp., Coh., Inf.: empathy, coherence, and informativity. Bold numbers indicate that the improvement of the method is statistically significant (paired t-test with p-value \textless 0.01). The scores where the student model outperforms the teacher are marked with *.}
	\begin{tabular}{cccccccccc|ccccccc}
		\hline
		\multirow{2}{*}{Model}  & \multirow{2}{*}{\#Params} & \multirow{2}{*}{Method} & \multicolumn{7}{c|}{MEDIC Dataset}       & \multicolumn{7}{c}{ED Dataset}            \\ \cline{4-17} 
		&                       &                         
		& Dist-1 & Dist-2 & Acc(\%) & F\textsubscript{BERT} & Emp. & Coh. & Inf. 
		& Dist-1 & Dist-2 & Acc(\%) & F\textsubscript{BERT} & Emp. & Coh. & Inf. \\ 
		\hline
		\multirow{7}{*}{\shortstack{Qwen2.5}}  & 7 B                     
		& Teacher & 13.19 & 39.54 & 67.29 & 88.65 & 3.69 & 3.76 & 3.77 
		& 13.39 & 40.95 & 44.96 & 87.32 & 3.75 & 3.72 & 3.75 \\ \cline{2-17} 
		& \multirow{6}{*}{3 B}    
		& SFT & 5.91    & 21.64 & 63.45 & 85.92 & 3.51 & 3.38 & 3.43 
		&6.64 & 22.15 & 41.28 & 85.57 & 3.65 & 3.36 & 3.44     \\
		&                         
		& SeqKD & 3.96  & 18.20 & 60.15 & 83.34 & 3.38 & 3.25 & 3.35     
		&4.20 & 18.77 & 36.10 & 83.06 & 3.40 & 3.24 & 3.35     \\
		&
		& SD &  8.62    & 26.86 & 65.21 & 86.50 & 3.61 & 3.58 & 3.52
		& 8.25& 27.15 & 42.30 & 85.65 & 3.60 & 3.53 & 3.50 \\
		&
		& KD &  3.67    & 17.83 & 61.56 & 83.84 & 3.40 & 3.29 & 3.30
		& 4.05 & 18.55 & 37.04 & 83.38 & 3.41 & 3.28 & 3.31 \\
		&                         
		& \textbf{PRIDE} 
		&\textbf{9.39} & \textbf{31.72} & \textbf{69.19*} & \textbf{88.76*} & \textbf{3.73*} & \textbf{3.75} & \textbf{3.66}     
		&\textbf{9.41} & \textbf{31.76} & \textbf{48.81*} & \textbf{87.94*}  & \textbf{3.79*} & \textbf{3.70} & \textbf{3.63}     \\
		&
		& w/o $\mathcal{P}$ & 5.46 & 21.15 & 63.12 & 85.65 & 3.48 & 3.35 & 3.40
		& 5.58 & 21.85 & 40.95 & 85.33 & 3.62 & 3.34 & 3.41     \\ \cline{2-17}
		\hline
		\multirow{7}{*}{LLaVA} & 7 B                    
		& Teacher & 11.55& 38.79 & 65.60 & 86.88 & 3.62 & 3.72 & 3.75     
		& 12.51& 39.42 & 43.84 & 85.80 & 3.70 & 3.71 & 3.76     \\ \cline{2-17}
		& \multirow{6}{*}{2 B}    
		& SFT &  5.13  & 22.16 & 62.06 & 85.26 & 3.43 & 3.35 & 3.43 
		& 6.03 & 22.42 & 40.11 & 85.01 & 3.50 & 3.32 & 3.44     \\
		&                         
		& SeqKD & 4.62 & 20.79 & 59.13 & 83.22 & 3.36 & 3.42 & 3.39 
		&4.07 & 20.19 & 35.56 & 82.61 & 3.38 & 3.42 & 3.37     \\
		&
		& SD &  7.42    & 24.12 & 63.51 & 85.90 & 3.57 & 3.55 & 3.41
		&6.54 & 24.45 & 40.82 & 84.83 & 3.59 & 3.53 & 3.38 \\
		&  
		& KD &  4.19   & 18.32 & 59.66 & 83.95 & 3.38 & 3.30 & 3.35
		&3.40 & 18.56 & 35.70 & 83.11 & 3.39 & 3.25 & 3.36 \\
		&                         
		& \textbf{PRIDE} & \textbf{7.92}& \textbf{28.25} & \textbf{66.82*} & \textbf{87.44*} &          \textbf{3.69*}  & \textbf{3.71}  & \textbf{3.47} 
		
		&\textbf{8.45} & \textbf{28.57} & \textbf{46.67} & \textbf{86.08*} & 
		\textbf{3.71*} & \textbf{3.70}  &  \textbf{3.43}    \\
		&
		& w/o $\mathcal{P}$ &4.93 & 21.88 & 61.85 & 84.95 & 3.41 & 3.34 & 3.40     
		& 5.86& 21.92 & 39.85 & 84.78 & 3.48 & 3.30 & 3.41     \\ \hline
		\multirow{13}{*}{Gemma3} & 12 B                    
		& Teacher &15.91 & 47.74 & 68.36 & 87.18 & 3.74 & 3.74 & 3.84     
		& 16.01& 47.53 & 46.50 & 85.97 & 3.76 & 3.73 & 3.82     \\ \cline{2-17} 
		& \multirow{6}{*}{4 B}
		& SFT  & 9.55   & 32.31 & 64.03 & 85.24 & 3.55 & 3.35 & 3.51      
		& 11.13& 32.56 & 41.15 & 84.16 & 3.65 & 3.46 & 3.51     \\
		&
		& SeqKD & 7.15  & 25.61 & 62.17 & 83.65 & 3.43 & 3.28 & 3.48     
		&6.47 & 25.48 & 38.40 & 82.91 & 3.46 & 3.24 & 3.47  \\
		&
		& SD  &  10.90   & 33.51 & 66.79 & 85.85 & 3.65 & 3.60 & 3.64
		&9.79 & 33.76 & 43.06 & 84.87 & 3.63 & 3.53 & 3.66 \\
		& 
		& KD  &  7.42   & 25.95 & 62.89 & 83.75 & 3.44 & 3.28 & 3.45
		&6.12 & 25.94 & 38.92 & 83.10 & 3.46 & 3.25 & 3.45 \\
		&                         
		& \textbf{PRIDE}& \textbf{12.14} & \textbf{39.65} & \textbf{69.55*} & \textbf{87.92*} & \textbf{3.78*} & \textbf{3.75*} & \textbf{3.78}   
		& \textbf{12.94}& \textbf{39.36} & \textbf{49.14*} & \textbf{86.45*} & \textbf{3.77*} & \textbf{3.61} & \textbf{3.78}    \\
		&
		& w/o $\mathcal{P}$ &10.21 & 31.95 & 63.85 & 84.98 & 3.52 & 3.33 & 3.49    
		&9.93 & 32.10 & 40.85 & 83.92 & 3.61 & 3.42 & 3.48     \\ \cline{2-17}
		& \multirow{6}{*}{1 B}    
		& SFT  & 4.05  & 18.24 & 61.17 & 84.76 & 3.41 & 3.26 & 3.35 
		& 3.95& 18.28 & 39.02 & 83.41 & 3.44 & 3.17 & 3.34     \\
		&                         
		& SeqKD & 2.54 & 16.55 & 60.02 & 83.13 & 3.38 & 3.11 & 3.30 
		&2.93 & 16.41 & 36.78 & 82.09 & 3.41 & 3.03 & 3.29     \\
		&
		& SD &  4.63   & 20.16 & 61.99 & 85.86 & 3.43 & 3.46 & 3.38
		& 5.46&19.95 & 39.36 & 84.35 & 3.45 & 3.37 & 3.37 \\
		&  
		& KD  &  3.47  & 16.16 & 60.19 & 83.98 & 3.38 & 3.23 & 3.29
		&2.20 &16.15 & 36.96 & 81.95 & 3.38 & 3.09 & 3.29 \\
		
		&                         
		& \textbf{PRIDE} & \textbf{6.18}  & \textbf{22.77} & \textbf{63.25}  & \textbf{87.30$^*$} & \textbf{3.49} & \textbf{3.62} & \textbf{3.44} 
		& \textbf{5.58}&\textbf{22.35} & \textbf{44.93} & \textbf{85.95} & \textbf{3.53} & \textbf{3.55}   &  \textbf{3.43}    \\
		&
		& w/o $\mathcal{P}$ & 3.58& 17.92 & 60.95 & 84.55 & 3.39 & 3.24 & 3.32     
		& 3.81 & 18.05 & 38.80 & 83.20 & 3.42 & 3.15 & 3.31     \\ \hline
	\end{tabular}
	\label{auto_eva}
\end{table*}

\begin{table*}[h]
	\small
	\centering
	\renewcommand{\arraystretch}{0.75}
	\caption{Ablation study of PRIDE components.}
	
		\begin{tabular}{l *{12}{c}}
			\toprule
			\multirow{3}{*}{\textbf{Ablation}} & \multicolumn{6}{c}{\textbf{MEDIC Dataset}} & \multicolumn{6}{c}{\textbf{ED Dataset}} \\
			\cmidrule(lr){2-7} \cmidrule(lr){8-13}
			& \multicolumn{3}{c}{Automatic Metrics} & \multicolumn{3}{c}{GPT-4o Eval} & \multicolumn{3}{c}{Automatic Metrics} & \multicolumn{3}{c}{GPT-4o Eval} \\
			\cmidrule(lr){2-4} \cmidrule(lr){5-7} \cmidrule(lr){8-10} \cmidrule(lr){11-13}
			& Dist-2 & Acc(\%) & F\textsubscript{BERT} & Emp. & Coh. & Inf. & Dist-2 & Acc(\%) & F\textsubscript{BERT} & Emp. & Coh. & Inf. \\
			\midrule
			
			w/o $L_{sim}$ & 25.21 & 63.95{\scriptsize$\downarrow$5.24} & 85.83 & 3.54 & 3.50 & 3.45 & 25.64 & 41.28{\scriptsize$\downarrow$7.53} & 85.11 & 3.55 & 3.51 & 3.48 \\
			w/o $L_{KD}$ & 24.13 & 63.52{\scriptsize$\downarrow$5.67} & 85.41 & 3.50 & 3.46 & 3.41 & 24.81 & 40.55{\scriptsize$\downarrow$8.26} & 84.72 & 3.51 & 3.47 & 3.44 \\
			w/o $L_{cls}$ & 30.85 & 65.58{\scriptsize$\downarrow$3.61} & 87.94 & 3.65 & 3.69 & 3.61 & 30.92 & 43.83{\scriptsize$\downarrow$4.98} & 86.60 & 3.68 & 3.65 & 3.59 \\
			w/ FKL & 27.85 & 66.40{\scriptsize$\downarrow$2.79} & 86.75 & 3.62 & 3.60 & 3.55 & 27.60 & 45.20{\scriptsize$\downarrow$3.61} & 86.30 & 3.64 & 3.58 & 3.55 \\
			w/ concat & 30.49 & 64.15{\scriptsize$\downarrow$5.04} & 88.13 & 3.67 & 3.70 & 3.61 & 30.55 & 42.15{\scriptsize$\downarrow$6.66} & 86.71 & 3.66 & 3.66 & 3.60 \\
			
			\bottomrule
		\end{tabular}
		
		\label{ablation}
\end{table*}
We considered several baselines in our main experiment. Specifically, \textbf{SFT} fine-tunes the student model directly supervised by the ground-truth responses, while \textbf{SeqKD} \cite{seqkd} fine-tunes the student on responses generated by the teacher. \textbf{SD} \cite{zhang2021self} adopts a self-distillation strategy, training the student on its own responses generated with access to privileged information. \textbf{KD} \cite{sanh2019distilbert} fine-tunes the student using the teacher's token-level probability distribution as supervision. \textbf{w/o $\mathcal{P}$} denotes our method without the privileged information. We also compare PRIDE with recent state-of-the-art methods on the ED datasets. The results are detailed in Table \ref{sota}.

\begin{table}[h]
	\normalsize
	\renewcommand{\tabcolsep}{2.0pt}
	\centering
	\caption{Comparison with state-of-the-art methods.}
	\begin{tabular}{cccccc}
		\hline
		\textbf{Methods} & \textbf{Acc(\%)} & \textbf{PPL $\downarrow$} & \textbf{Dist-1} & \textbf{Dist-2} & F\textsubscript{BERT} \\ \hline
		MoEL\cite{moel}  & 38.35 & 32.20 & 0.47 & 2.05 & 80.72      \\
		MIME\cite{mime}  & 37.33 & 29.60 & 0.41 & 2.62  & 81.35      \\
		EmpDG\cite{empdg} & 34.31 & 37.29 & 0.46 & 2.02 & 81.85     \\
		CEM\cite{cem}   & 39.11 & 36.11 & 0.66 & 2.99   & 82.20     \\ 
		SEEK\cite{seek} & 41.85 & 37.09 & 0.73 & 3.23  & 81.98     \\ 
		EmpSOA\cite{empsoa} & 35.02 & 48.32 & 0.71 & 3.96 & 82.13     \\ 
		CASE\cite{case} & 40.20 & 35.37 & 0.74 & 4.01 & 82.24     \\ 
		Empatheia\cite{zhang2025towards} & 48.51 & 11.67 & 2.69 & 14.76 & 86.87     \\ \hline 
		PRIDE\textsubscript{Qwen} & \textbf{48.81} & \textbf{7.92} & \textbf{9.70} & \textbf{31.76} & \textbf{87.94} \\ \hline
	\end{tabular}
	
	\label{sota}
\end{table}

\subsection{Evaluation Metrics}
\subsubsection{Automatic Evaluation}
Following previous works \cite{cem,seek}, we employed PPL, Dist (Dist-1 and Dist-2) \cite{li-etal-2016-diversity} and Acc to evaluate the generated responses in quality, diversity and average emotion recognition accuracy. We also adopted BERTScore ($F_{BERT}$) \citep{bertscore} to measure the semantic similarity between the generated responses and the ground truth. The average accuracy of emotion recognition reflects the quality of affective empathy in responses, and BERTScore evaluates the level of cognitive empathy in linguistic content \cite{yan-etal-2024-talk}. We also employed GPT-4o as a simulated evaluator. The GPT-4o evaluation follows three aspects: (1) \textbf{Empathy}: The response demonstrates comprehension of the speaker's emotional state and situations; (2) \textbf{Coherence}: The response coherently aligns with the dialogue context; (3) \textbf{Informativity}: The response provides informative content. Ratings range from 1 (unacceptable) to 5 (excellent).

\subsubsection{Human Evaluation}
While automatic metrics are important, they cannot fully replace human perception of empathy. Therefore, we also conducted human evaluations. We randomly sampled 400 response pairs from the testing datasets and engaged 11 third-party researchers with master's degrees in psychology to evaluate the responses. The evaluations are also based on the aspects of empathy, coherence and informativity. Furthermore, we conducted aspect-based pairwise comparisons. Concretely, for a given context, we randomly paired our model's response with one from another model and asked annotators to choose \textit{Win} or \textit{Lose} to indicate their preference. \textit{Tie} is also allowed, but the annotators were encouraged to choose one. We measured inter-annotator agreement using Fleiss' kappa \cite{kappa}, yielding values between 0.43 and 0.51, which indicates moderate agreement. To ensure a focused analysis, this evaluation primarily compares the responses generated by Qwen2.5-VL student model (3B) against its corresponding teacher and other baseline models. The detailed guidelines provided to the annotators are provided in Appendix.

\section{Results Analysis}
\subsection{Comparison with Baselines and SOTA}
Table \ref{auto_eva} presents a comparison of our method with baselines on the two datasets. The results show that, regardless of the model architecture or dataset, PRIDE outperforms baselines across all automatic evaluation metrics.

Notably, the student model trained with PRIDE even surpasses its teacher model in some cases. For example, on the MEDIC dataset, the Qwen2.5-VL (3B) student model achieves an emotion recognition accuracy of 69.19\%, exceeding the teacher's 67.29\%. Similarly, on the ED dataset, its accuracy exceeds the teacher's by 3.85\%, while its F\textsubscript{BERT} score also shows an improvement of 0.62. Notably, on the MEDIC dataset, even the much smaller Gemma3 (1B) student model outperforms its teacher (12B) on semantic similarity metrics. The 1B student's F\textsubscript{BERT} score is higher than the 12B teacher's, exceeding it by 0.12 points. These results provides strong evidence for the advantage of our method. 
The multi-source attention mechanism provides the student with emotional and situational context from the privileged information, which is reflected in the improved Acc. This understanding is then internalized through the dual-alignment losses, which aligns the student's semantic space with the teacher's at both feature and logit levels, explaining the significant gains in semantic similarity.
However, we must emphasize that the student-surpasses-teacher phenomenon arises under specific conditions: the student model internalizes knowledge from privileged information during its training, whereas the teacher model does not have access to this information during inference. Therefore, this result should be interpreted not as evidence that the student's architecture is inherently superior, but rather as a strong validation of the effectiveness of the PRIDE training paradigm itself. We attribute this performance gain to the superior efficiency of the multi-source attention mechanism in processing privileged information compared to the teacher's standard approach. While the teacher model handles privileged information via simple concatenation within a single context window (standard self-attention), this can lead to attentional dispersion where critical emotional cues are diluted by the lengthy dialogue history. In contrast, our student model disentangles dialogue context from privileged insights. This structural separation allows the student to focus on the high-level empathetic reasoning embedded in the privileged information more effectively. By dynamically fusing these distinct signals through the gating mechanism, the student constructs a more robust latent representation that persists even when the privileged input is removed at inference.

Furthermore, the comparison with the w/o P (without privileged information) highlights the essential role of privileged information in PRIDE. In all cases, removing privileged information leads to a substantial performance degradation, proving our hypothesis: privileged information can enhance the quality of empathetic dialogue generation.
In Table \ref{sota}, we compare our student model (using Qwen2.5-VL 3B) with existing state-of-the-art methods in empathetic dialogue domain on the ED dataset. The results show that PRIDE surpasses all existing SOTA methods on all metrics. Our PPL of 7.92 is lower than all competing methods, indicating that the generated responses are more linguistically fluent. Dist-1 and Dist-2 scores (9.70 and 31.76) far exceed Empatheia \cite{zhang2025towards} (2.69 and 14.76), which signifies a substantial improvement in generating diverse responses.

\subsection{Ablation Study}
To validate the effectiveness of each component within PRIDE, we conducted ablation studies in Table \ref{ablation}. The experiments were performed on the Qwen2.5-VL (3B) model. The results show that removing either alignment loss (w/o $L_{sim}$, w/o $L_{KD}$) leads to a significant drop in all metrics. This confirms the necessity of the dual-alignment strategy. We also investigated the impact of the divergence function choice, a key component of our distillation process.
Replacing our proposed RKLD with the standard Forward Kullback-Leibler divergence (w/ Forward KL) resulted in a noticeable performance drop.
Specifically, the Dist-2 score on the MEDIC dataset decreased from 31.72 to 27.85.
This indicates that the mean-seeking nature of Forward KL tends to average out the teacher's distinct predictions, leading to more generic responses.
In contrast, the mode-seeking property of RKLD forces the student to capture the teacher's confident and specific empathetic expressions, thereby preserving diversity and specificity. 
After removing $L_{cls}$, the model's performance also declined, especially in emotion recognition accuracy (Acc). This suggests that guiding the student model to recognize emotions enhances its affective empathy, which in turn serves to generate empathetic responses. We also tested the effectiveness of the multi-source attention mechanism. To test this, we replaced the mechanism with a simple concatenation approach that fuses the dialogue context and privileged information (w/ concat). The results show that the concatenation approach is inferior to our proposed method, which substantiates the superiority of our multi-source attention mechanism. Our method extracts relevant context from the privileged information, thus avoiding the pitfalls of naive concatenation.

\begin{figure}[h]
	\centering
	\includegraphics[width=0.95\linewidth]{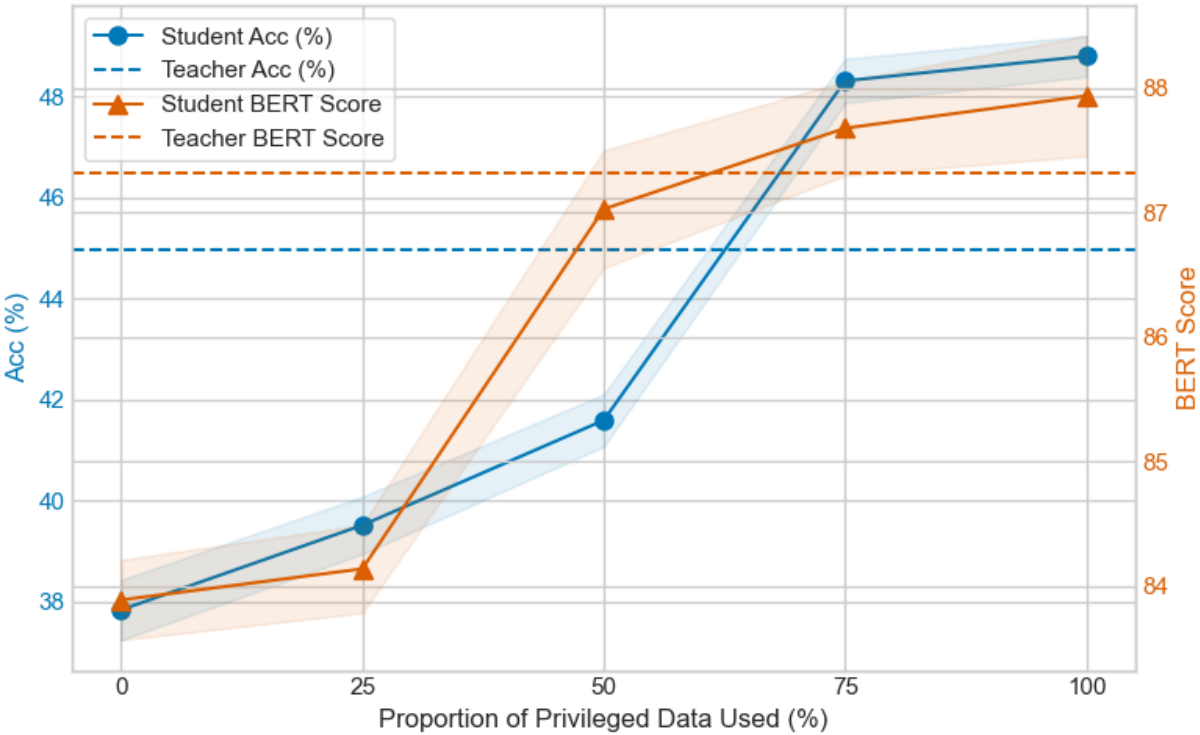}
	\caption{Impact of Privileged Information Proportion.}
	\label{line}
\end{figure}
\subsection{Privileged Information Proportion Analysis}
We analyzed how the proportion of privileged information affects performance, as shown in Figure \ref{line}, more privileged information leads to better results. Acc improves steadily with more data. The most significant gain occurs when increasing the data proportion from 50\% to 75\%. The student's accuracy surpasses the fixed teacher accuracy level after using 75\% of the privileged information. The F\textsubscript{BERT} score also shows a strong upward trend. Its largest performance jump happens between the 25\% and 50\% data points. The student's F\textsubscript{BERT} score surpasses the teacher's score very early, after using just 50\% of the data. This analysis confirms that privileged information is the core driver of performance in PRIDE. It also highlights the method's data efficiency, as the student model can outperform its teacher using 75\% of the total privileged information.
\begin{figure}[h]
	\centering
	\includegraphics[width=0.95\linewidth]{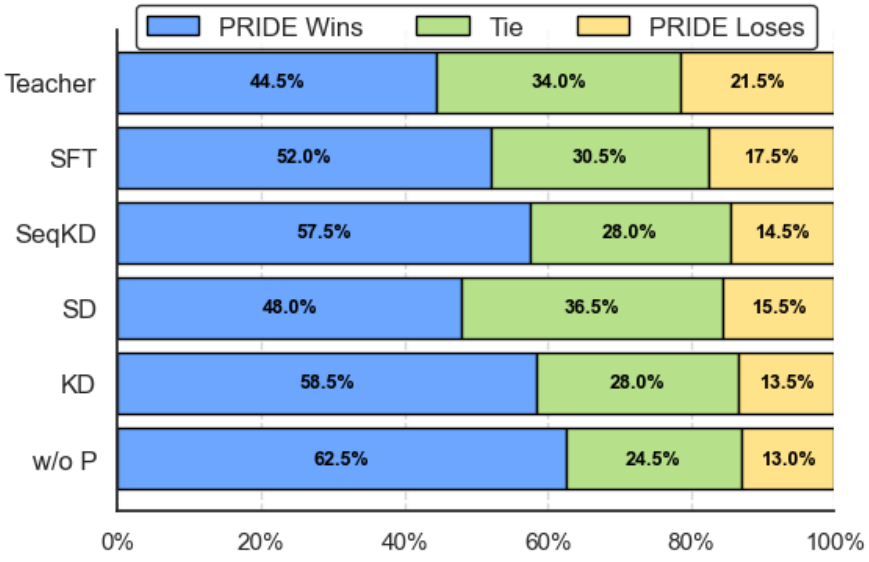}
	\caption{Results of pairwise human evaluation.}
	\label{abtest}
\end{figure}

\subsection{Human Preference and Quality Analysis}
\begin{figure}[h]
	\centering
	\includegraphics[width=0.95\linewidth]{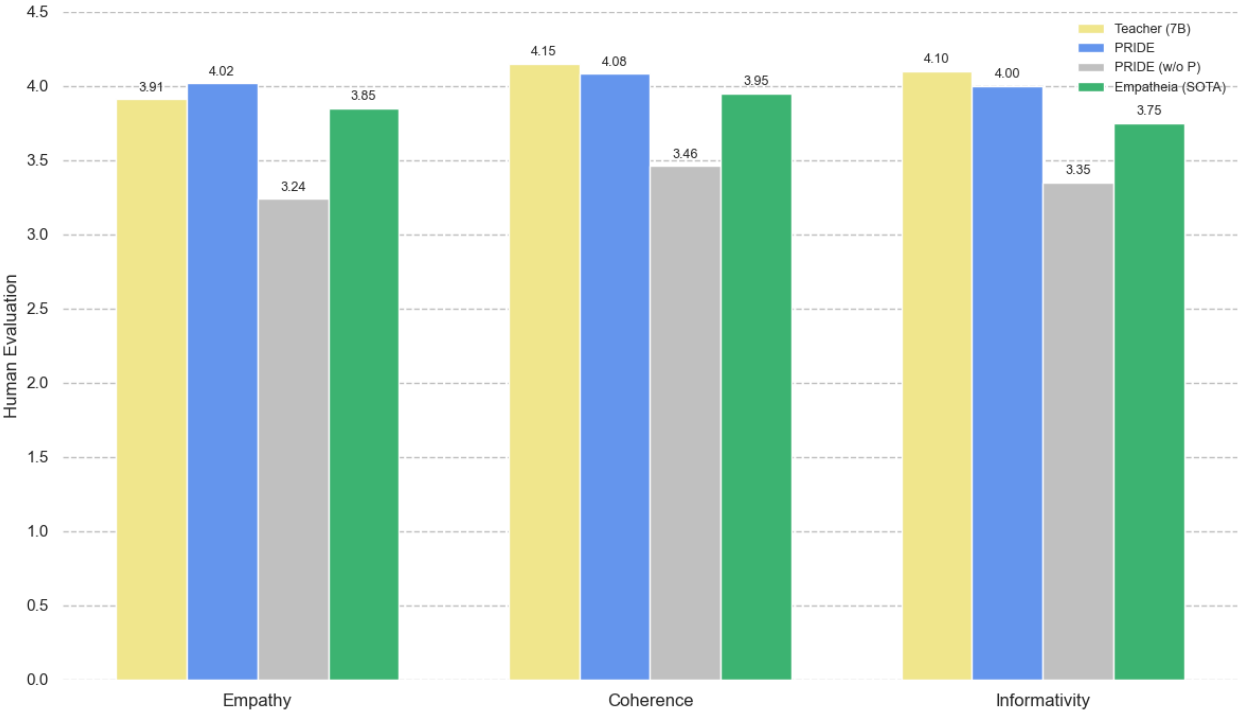}
	\caption{Human evaluation scores.}
	\label{human}
\end{figure}

Figure \ref{abtest} and Figure \ref{human} collectively provide evidence of PRIDE's superiority from a human-centered perspective. In pairwise preference evaluations (Figure \ref{abtest}), PRIDE achieves the highest win rates across all comparisons, with a particularly notable 44.5\% win rate against the teacher model. This confirms that the student model is not only capable of retaining the teacher's empathetic abilities but can surpass them in generating responses through the integration of privileged information and dual-alignment training.
As Figure \ref{human} shows, PRIDE receives the highest score in the empathy dimension. 
While PRIDE does not surpass the teacher model or the latest state-of-the-art method (Empatheia) across every metric, it achieves competitive or superior performance.
Overall, the results of human evaluation are consistent with the automatic metrics, and validate the practical value of our method in enhancing the quality of empathetic dialogue.

\subsection{Qualitative Analysis}
\begin{figure}[h!]
	\centering
	\includegraphics[width=1.02\linewidth]{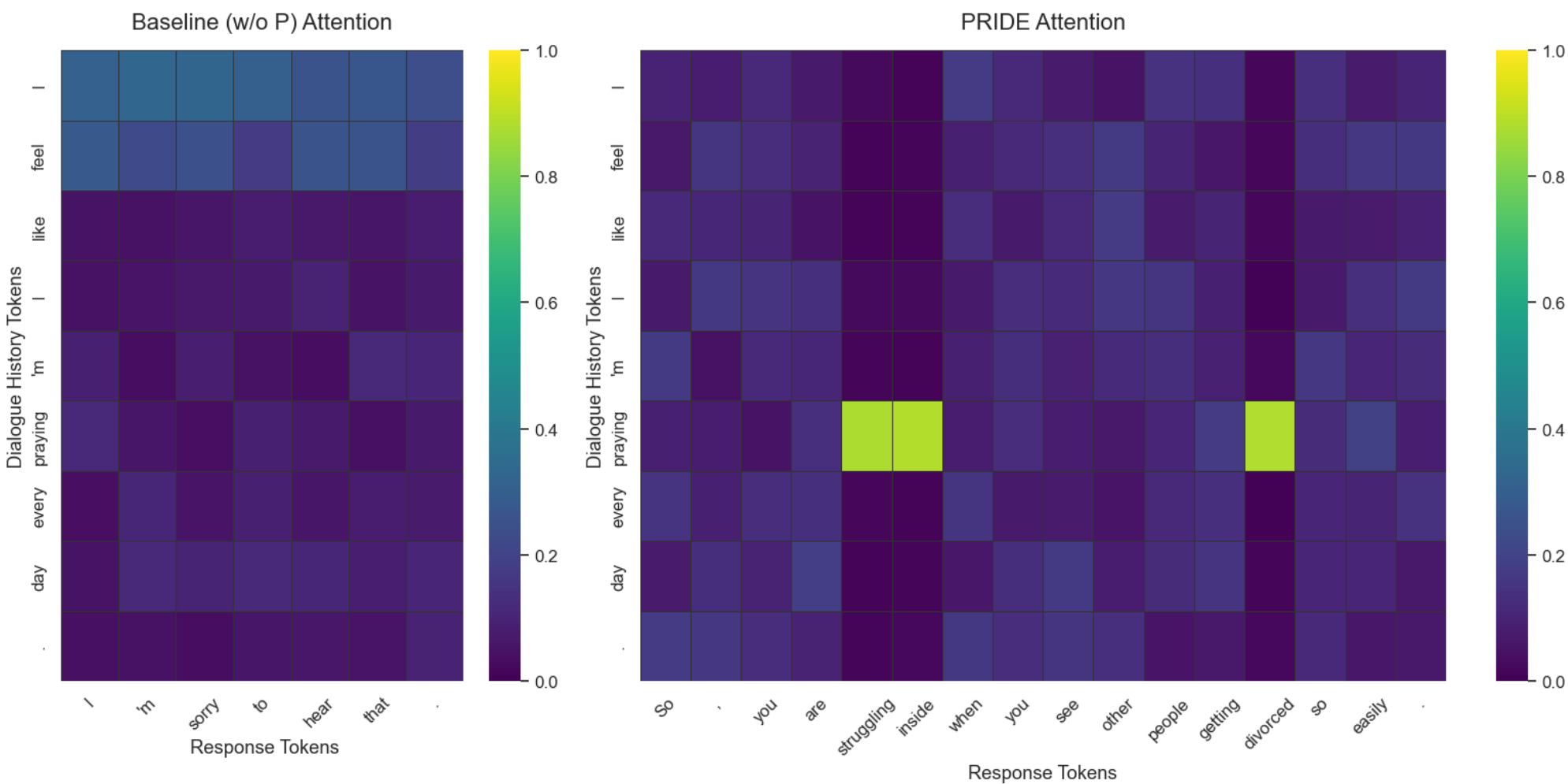}
	\caption{Multi-source attention visualization.}
	\label{heat}
\end{figure}

To investigate how the student model leverages privileged information during inference, where such information is physically absent, we visualized the attention weights of the PRIDE student model compared to the baseline (w/o $\mathcal{P}$).
Figure \ref{heat} presents a case from the MEDIC dataset where the user says: \textit{``I feel like I'm praying every day...''}.

As shown in Figure \ref{heat}, the PRIDE student exhibits a distinct attention pattern that aligns with the implicit psychological context.
Specifically, when generating the context-specific keywords ``struggling'' and ``divorced'', the model places significantly higher attention weights on the history token ``praying''.
This behavior is remarkable because the connection between ``praying'' and ``fear of divorce'' was only explicit in the privileged expert analysis (which is unavailable during inference).
The high attention weight confirms that the student has internalized this reasoning: it treats ``praying'' as a semantic anchor to infer the user's hidden anxiety, thereby generating the profound response: \textit{``So, you are struggling inside when you see other people getting divorced...''}.

In contrast, the \textbf{w/o $\mathcal{P}$} baseline displays a diffuse attention distribution, mostly focusing on functional words like ``I'' and ``feel''.
Lacking the internalized privileged insight, thus the baseline fails to capture the latent intent and produces a generic response: \textit{``I'm sorry to hear that''}.
This comparison qualitatively validates that PRIDE does not merely mimic surface patterns but effectively transfers deep empathetic reasoning capabilities.

\subsection{Efficiency Analysis}
To quantify the practical benefits of our distillation method, we compared the inference efficiency of the teacher and student models. We measured the memory usage and inference speed (tokens per second) on a single NVIDIA RTX 4090 (24GB) GPU with a batch size of 1.

\begin{table}[h]
	\centering
	\small
	\renewcommand{\arraystretch}{1.1}
	\setlength{\tabcolsep}{2pt}
	\caption{Efficiency Comparison between Teacher and Student Models.}
	\begin{tabular}{l c c c c}
		\toprule
		\textbf{Model} & \textbf{Params} & \textbf{Memory $\downarrow$} & \textbf{Speed (tok/s) $\uparrow$} & \textbf{Speedup} \\
		\midrule
		\multicolumn{5}{l}{\textit{Qwen2.5}} \\
		Teacher & 7B & 15.2 GB & 38.5 & 1.00$\times$ \\
		Student & \textbf{3B} & \textbf{7.1} GB & \textbf{79.2} & \textbf{2.06$\times$} \\
		\midrule
		\multicolumn{5}{l}{\textit{LLaVA}} \\
		Teacher & 7B & 14.8 GB & 40.1 & 1.00$\times$ \\
		Student & \textbf{2B} & \textbf{4.9} GB & \textbf{115.6} & \textbf{2.88$\times$} \\
		\midrule
		\multicolumn{5}{l}{\textit{Gemma3}} \\
		Teacher & 12B & 23.4 GB & 23.5 & 1.00$\times$ \\
		Student & \textbf{4B} & 9.2 GB & 65.8 & 2.80$\times$ \\
		Student & \textbf{1B} & \textbf{2.8} GB & \textbf{142.5} & \textbf{6.06$\times$} \\
		\bottomrule
	\end{tabular}
	\label{efficiency_table}
\end{table}
As shown in the table, the student models achieve significant efficiency gains, making them suitable for consumer-grade hardware.
The 12B teacher model nearly exhausts the 24GB VRAM limit of the RTX 4090, creating a bottleneck for deployment.
In contrast, the Gemma3 (1B) student model requires only 2.8 GB of VRAM (an 88.0\% reduction and achieves a remarkable speed of 142.5 tokens/s (a 6.06$\times$ speedup).
Similarly, the Qwen2.5-VL (3B) student reduces memory usage by 53.3\% compared to its teacher and doubles the inference throughput.
These results confirm that PRIDE effectively transfers knowledge to lightweight models, enabling high-performance empathetic dialogue generation on standard consumer GPUs with low latency.
\section{Conclusion}
In this paper, we propose PRIDE, a privileged information-enhanced knowledge distillation method for empathetic dialogue generation. Experiments on both multi-modal and text-only datasets demonstrate that PRIDE outperforms the baselines and state-of-the-art methods. A notable finding is that the privileged information we used allows student models to even surpass their larger teachers on certain metrics. This result strongly validates the effectiveness of our method.

\bibliographystyle{IEEEtran}
\bibliography{tac}

@inproceedings{empathyear,
	title = "{E}mpathy{E}ar: An Open-source Avatar Multimodal Empathetic Chatbot",
	author = "Fei, Hao  and
	Zhang, Han  and
	Wang, Bin  and
	Liao, Lizi  and
	Liu, Qian  and
	Cambria, Erik",
	editor = "Cao, Yixin  and
	Feng, Yang  and
	Xiong, Deyi",
	booktitle = "Proceedings of the 62nd Annual Meeting of the Association for Computational Linguistics (Volume 3: System Demonstrations)",
	month = aug,
	year = "2024",
	address = "Bangkok, Thailand",
	publisher = "Association for Computational Linguistics",
	pages = "61--71",
}

@book{damon2006handbook,
	title={Handbook of child psychology, social, emotional, and personality development},
	author={Damon, William and Lerner, Richard M and Eisenberg, Nancy},
	year={2006},
	publisher={John Wiley \& Sons}
}

@inproceedings{li-etal-2016-diversity,
    title = "A Diversity-Promoting Objective Function for Neural Conversation Models",
    author = "Li, Jiwei  and
      Galley, Michel  and
      Brockett, Chris  and
      Gao, Jianfeng  and
      Dolan, Bill",
    editor = "Knight, Kevin  and
      Nenkova, Ani  and
      Rambow, Owen",
    booktitle = "Proceedings of the 2016 Conference of the North {A}merican Chapter of the Association for Computational Linguistics: Human Language Technologies",
    month = jun,
    year = "2016",
    address = "San Diego, California",
    publisher = "Association for Computational Linguistics",
    url = "https://aclanthology.org/N16-1014/",
    doi = "10.18653/v1/N16-1014",
    pages = "110--119"
}

@article{vapnik2015learning,
  title={Learning using privileged information: Similarity control and knowledge transfer.},
  author={Vapnik, Vladimir and Izmailov, Rauf and others},
  journal={J. Mach. Learn. Res.},
  volume={16},
  number={1},
  pages={2023--2049},
  year={2015}
}

@article{aff-cog,
	title={Measuring individual differences in empathy: Evidence for a multidimensional approach.},
	author={Davis, Mark H},
	journal={Journal of personality and social psychology},
	volume={44},
	number={1},
	pages={113},
	year={1983},
	publisher={American Psychological Association}
}

@misc{zhao2023chatgptequip,
	title={Is ChatGPT Equipped with Emotional Dialogue Capabilities?}, 
	author={Weixiang Zhao and Yanyan Zhao and Xin Lu and Shilong Wang and Yanpeng Tong and Bing Qin},
	year={2023},
	eprint={2304.09582},
	archivePrefix={arXiv},
	primaryClass={cs.CL},
}

@inproceedings{mmd,
	author = {Gretton, Arthur and Borgwardt, Karsten and Rasch, Malte and Sch\"{o}lkopf, Bernhard and Smola, Alex},
	booktitle = {Advances in Neural Information Processing Systems},
	editor = {B. Sch\"{o}lkopf and J. Platt and T. Hoffman},
	pages = {},
	publisher = {MIT Press},
	title = {A Kernel Method for the Two-Sample-Problem},
	url = {https://proceedings.neurips.cc/paper_files/paper/2006/file/e9fb2eda3d9c55a0d89c98d6c54b5b3e-Paper.pdf},
	volume = {19},
	year = {2006}
}

@article{compress,   title={On-Policy Distillation of Language Models: Learning from Self-Generated Mistakes},  author={Agarwal, Rishabh and Vieillard, Nino and Zhou, Yongchao and Stanczyk, Piotr and Ramos, Sabela and Geist, Matthieu and Bachem, Olivier},  year={2024},  month={Jan},  language={en-US}  }

@misc{selfreward,
	title={Self-Rewarding Language Models}, 
	author={Weizhe Yuan and Richard Yuanzhe Pang and Kyunghyun Cho and Xian Li and Sainbayar Sukhbaatar and Jing Xu and Jason Weston},
	year={2025},
	eprint={2401.10020},
	archivePrefix={arXiv},
	primaryClass={cs.CL},
	url={https://arxiv.org/abs/2401.10020}, 
}

@article{rusu2015policy,
	title={Policy distillation},
	author={Rusu, Andrei A and Colmenarejo, Sergio Gomez and Gulcehre, Caglar and Desjardins, Guillaume and Kirkpatrick, James and Pascanu, Razvan and Mnih, Volodymyr and Kavukcuoglu, Koray and Hadsell, Raia},
	journal={arXiv preprint arXiv:1511.06295},
	year={2015}
}

@article{gou2021knowledge,
	title={Knowledge distillation: A survey},
	author={Gou, Jianping and Yu, Baosheng and Maybank, Stephen J and Tao, Dacheng},
	journal={International Journal of Computer Vision},
	volume={129},
	number={6},
	pages={1789--1819},
	year={2021},
	publisher={Springer}
}

@article{seqkd,
  title={LightPAFF: A two-stage distillation framework for pre-training and fine-tuning},
  author={Song, Kaitao and Sun, Hao and Tan, Xu and Qin, Tao and Lu, Jianfeng and Liu, Hongzhi and Liu, Tie-Yan},
  journal={arXiv preprint arXiv:2004.12817},
  year={2020}
}

@article{Qwen2.5-VL,
  title={Qwen2.5-VL Technical Report},
  author={Bai, Shuai and Chen, Keqin and Liu, Xuejing and Wang, Jialin and Ge, Wenbin and Song, Sibo and Dang, Kai and Wang, Peng and Wang, Shijie and Tang, Jun and Zhong, Humen and Zhu, Yuanzhi and Yang, Mingkun and Li, Zhaohai and Wan, Jianqiang and Wang, Pengfei and Ding, Wei and Fu, Zheren and Xu, Yiheng and Ye, Jiabo and Zhang, Xi and Xie, Tianbao and Cheng, Zesen and Zhang, Hang and Yang, Zhibo and Xu, Haiyang and Lin, Junyang},
  journal={arXiv preprint arXiv:2502.13923},
  year={2025}
}

@article{gemma_2025,
    title={Gemma 3},
    url={https://goo.gle/Gemma3Report},
    publisher={Kaggle},
    author={Gemma Team},
    year={2025}
}

@misc{hinck2024llavagemma,
      title={LLaVA-Gemma: Accelerating Multimodal Foundation Models with a Compact Language Model}, 
      author={Musashi Hinck and Matthew L. Olson and David Cobbley and Shao-Yen Tseng and Vasudev Lal},
      year={2024},
      eprint={2404.01331},
      url={https://arxiv.org/abs/2404.01331},
      archivePrefix={arXiv},
      primaryClass={cs.CL}
}

@misc{gu2024minillm,
	title={MiniLLM: Knowledge Distillation of Large Language Models}, 
	author={Yuxian Gu and Li Dong and Furu Wei and Minlie Huang},
	year={2024},
	eprint={2306.08543},
	archivePrefix={arXiv},
	primaryClass={cs.CL},
	url={https://arxiv.org/abs/2306.08543}, 
}

@misc{xu2024surveykd,
	title={A Survey on Knowledge Distillation of Large Language Models}, 
	author={Xiaohan Xu and Ming Li and Chongyang Tao and Tao Shen and Reynold Cheng and Jinyang Li and Can Xu and Dacheng Tao and Tianyi Zhou},
	year={2024},
	eprint={2402.13116},
	archivePrefix={arXiv},
	primaryClass={cs.CL},
	url={https://arxiv.org/abs/2402.13116}, 
}

@inproceedings{cot,
	author = {Wei, Jason and Wang, Xuezhi and Schuurmans, Dale and Bosma, Maarten and ichter, brian and Xia, Fei and Chi, Ed and Le, Quoc V and Zhou, Denny},
	booktitle = {Advances in Neural Information Processing Systems},
	editor = {S. Koyejo and S. Mohamed and A. Agarwal and D. Belgrave and K. Cho and A. Oh},
	pages = {24824--24837},
	publisher = {Curran Associates, Inc.},
	title = {Chain-of-Thought Prompting Elicits Reasoning in Large Language Models},
	url = {https://proceedings.neurips.cc/paper_files/paper/2022/file/9d5609613524ecf4f15af0f7b31abca4-Paper-Conference.pdf},
	volume = {35},
	year = {2022}
}

@inproceedings{yang-iterative,
	title = "An Iterative Associative Memory Model for Empathetic Response Generation",
	author = "Yang, Zhou  and
	Ren, Zhaochun  and
	Yufeng, Wang  and
	Sun, Haizhou  and
	Chen, Chao  and
	Zhu, Xiaofei  and
	Liao, Xiangwen",
	editor = "Ku, Lun-Wei  and
	Martins, Andre  and
	Srikumar, Vivek",
	booktitle = "Proceedings of the 62nd Annual Meeting of the Association for Computational Linguistics (Volume 1: Long Papers)",
	month = aug,
	year = "2024",
	address = "Bangkok, Thailand",
	publisher = "Association for Computational Linguistics",
	pages = "3081--3092",
}

@article{kappa,
	title={The equivalence of weighted kappa and the intraclass correlation coefficient as measures of reliability},
	author={Fleiss, Joseph L and Cohen, Jacob},
	journal={Educational and psychological measurement},
	volume={33},
	number={3},
	pages={613--619},
	year={1973},
	publisher={Sage Publications Sage CA: Thousand Oaks, CA}
}

@misc{hinton2015distilling,
	title={Distilling the Knowledge in a Neural Network}, 
	author={Geoffrey Hinton and Oriol Vinyals and Jeff Dean},
	year={2015},
	eprint={1503.02531},
	archivePrefix={arXiv},
	primaryClass={stat.ML},
}

@inproceedings{bertscore,
	author       = {Tianyi Zhang and
	Varsha Kishore and
	Felix Wu and
	Kilian Q. Weinberger and
	Yoav Artzi},
	title        = {BERTScore: Evaluating Text Generation with {BERT}},
	booktitle    = {8th International Conference on Learning Representations, {ICLR} 2020,
	Addis Ababa, Ethiopia, April 26-30, 2020},
	publisher    = {OpenReview.net},
	year         = {2020},
	url          = {https://openreview.net/forum?id=SkeHuCVFDr},
	bibsource    = {dblp computer science bibliography, https://dblp.org}
}

@article{zhang2021self,
  title={Self-distillation: Towards efficient and compact neural networks},
  author={Zhang, Linfeng and Bao, Chenglong and Ma, Kaisheng},
  journal={IEEE Transactions on Pattern Analysis and Machine Intelligence},
  volume={44},
  number={8},
  pages={4388--4403},
  year={2021},
  publisher={IEEE}
}

@inproceedings{yan-etal-2024-talk,
	title = "Talk With Human-like Agents: Empathetic Dialogue Through Perceptible Acoustic Reception and Reaction",
	author = "Yan, Haoqiu  and
	Zhu, Yongxin  and
	Zheng, Kai  and
	Liu, Bing  and
	Cao, Haoyu  and
	Jiang, Deqiang  and
	Xu, Linli",
	editor = "Ku, Lun-Wei  and
	Martins, Andre  and
	Srikumar, Vivek",
	booktitle = "Proceedings of the 62nd Annual Meeting of the Association for Computational Linguistics (Volume 1: Long Papers)",
	month = aug,
	year = "2024",
	address = "Bangkok, Thailand",
	publisher = "Association for Computational Linguistics",
	url = "https://aclanthology.org/2024.acl-long.801",
	pages = "15009--15022",
}

@inproceedings{stickerconv,
	title = "{STICKERCONV}: Generating Multimodal Empathetic Responses from Scratch",
	author = "Zhang, Yiqun  and
	Kong, Fanheng  and
	Wang, Peidong  and
	Sun, Shuang  and
	SWangLing, SWangLing  and
	Feng, Shi  and
	Wang, Daling  and
	Zhang, Yifei  and
	Song, Kaisong",
	editor = "Ku, Lun-Wei  and
	Martins, Andre  and
	Srikumar, Vivek",
	booktitle = "Proceedings of the 62nd Annual Meeting of the Association for Computational Linguistics (Volume 1: Long Papers)",
	month = aug,
	year = "2024",
	address = "Bangkok, Thailand",
	publisher = "Association for Computational Linguistics",
	url = "https://aclanthology.org/2024.acl-long.417",
	pages = "7707--7733",
}

@inproceedings{medic,
	author = {Zhu, Zhouan and Li, Chenguang and Pan, Jicai and Li, Xin and Xiao, Yufei and Chang, Yanan and Zheng, Feiyi and Wang, Shangfei},
	title = {MEDIC: A Multimodal Empathy Dataset in Counseling},
	year = {2023},
	isbn = {9798400701085},
	publisher = {Association for Computing Machinery},
	address = {New York, NY, USA},
	url = {https://doi.org/10.1145/3581783.3612346},
	doi = {10.1145/3581783.3612346},
	booktitle = {Proceedings of the 31st ACM International Conference on Multimedia},
	pages = {6054–6062},
	numpages = {9},
	location = {Ottawa ON, Canada},
	series = {MM '23}
}

@ARTICLE{survey2023,
	author={Raamkumar, Aravind Sesagiri and Yang, Yinping},
	journal={IEEE Transactions on Affective Computing}, 
	title={Empathetic Conversational Systems: A Review of Current Advances, Gaps, and Opportunities}, 
	year={2023},
	volume={14},
	number={4},
	pages={2722-2739},
	doi={10.1109/TAFFC.2022.3226693}}

@article{macarov,
	title={Empathy: The charismatic chimera},
	author={Macarov and David},
	journal={Journal of Education for Social Work},
	volume={14},
	number={3},
	pages={86--92},
	year={1978},
	publisher={Taylor \& Francis}
}

@inproceedings{empatheticdialogues,
	title = "Towards Empathetic Open-domain Conversation Models: A New Benchmark and Dataset",
	author = "Rashkin, Hannah  and
	Smith, Eric Michael  and
	Li, Margaret  and
	Boureau, Y-Lan",
	booktitle = "Proceedings of the 57th Annual Meeting of the Association for Computational Linguistics",
	month = jul,
	year = "2019",
	address = "Florence, Italy",
	publisher = "Association for Computational Linguistics",
	pages = "5370--5381",
}

@inproceedings{moel,
	title = "{M}o{EL}: Mixture of Empathetic Listeners",
	author = "Lin, Zhaojiang  and
	Madotto, Andrea  and
	Shin, Jamin  and
	Xu, Peng  and
	Fung, Pascale",
	booktitle = "Proceedings of the 2019 Conference on Empirical Methods in Natural Language Processing and the 9th International Joint Conference on Natural Language Processing (EMNLP-IJCNLP)",
	month = nov,
	year = "2019",
	address = "Hong Kong, China",
	publisher = "Association for Computational Linguistics",
	pages = "121--132",
}

@inproceedings{mime,
	title={MIME: MIMicking Emotions for Empathetic Response Generation},
	author={Ghosal, Debanjan and Majumder, Bodhisattwa Prasad and Poria, Soujanya and Gelbukh, Alexander and Cambria, Erik},
	booktitle={Proceedings of the 2020 Conference on Empirical Methods in Natural Language Processing (EMNLP)},
	pages={7645--7655},
	year={2020}
}

@inproceedings{seek,
	title = "Empathetic Dialogue Generation via Sensitive Emotion Recognition and Sensible Knowledge Selection",
	author = "Wang, Lanrui  and
	Li, Jiangnan  and
	Lin, Zheng  and
	Meng, Fandong  and
	Yang, Chenxu  and
	Wang, Weiping  and
	Zhou, Jie",
	booktitle = "Findings of the Association for Computational Linguistics: EMNLP 2022",
	month = dec,
	year = "2022",
	address = "Abu Dhabi, United Arab Emirates",
	publisher = "Association for Computational Linguistics",
	pages = "4634--4645",
}

@inproceedings{case,
	title = "{CASE}: Aligning Coarse-to-Fine Cognition and Affection for Empathetic Response Generation",
	author = "Zhou, Jinfeng  and
	Zheng, Chujie  and
	Wang, Bo  and
	Zhang, Zheng  and
	Huang, Minlie",
	editor = "Rogers, Anna  and
	Boyd-Graber, Jordan  and
	Okazaki, Naoaki",
	booktitle = "Proceedings of the 61st Annual Meeting of the Association for Computational Linguistics (Volume 1: Long Papers)",
	month = jul,
	year = "2023",
	address = "Toronto, Canada",
	publisher = "Association for Computational Linguistics",
	url = "https://aclanthology.org/2023.acl-long.457",
	doi = "10.18653/v1/2023.acl-long.457",
	pages = "8223--8237",
}

@inproceedings{empdg,
	title = "{E}mp{DG}: Multi-resolution Interactive Empathetic Dialogue Generation",
	author = "Li, Qintong  and
	Chen, Hongshen  and
	Ren, Zhaochun  and
	Ren, Pengjie  and
	Tu, Zhaopeng  and
	Chen, Zhumin",
	booktitle = "Proceedings of the 28th International Conference on Computational Linguistics",
	month = dec,
	year = "2020",
	address = "Barcelona, Spain (Online)",
	publisher = "International Committee on Computational Linguistics",
	pages = "4454--4466",
}

@inproceedings{cem,
	title={Cem: Commonsense-aware empathetic response generation},
	author={Sabour, Sahand and Zheng, Chujie and Huang, Minlie and Huang, Minlie},
	booktitle={Proceedings of the AAAI Conference on Artificial Intelligence},
	volume={36},
	number={10},
	pages={11229--11237},
	year={2022}
}

@inproceedings{comet,
	title = "{COMET}: Commonsense Transformers for Automatic Knowledge Graph Construction",
	author = "Bosselut, Antoine  and
	Rashkin, Hannah  and
	Sap, Maarten  and
	Malaviya, Chaitanya  and
	Celikyilmaz, Asli  and
	Choi, Yejin",
	booktitle = "Proceedings of the 57th Annual Meeting of the Association for Computational Linguistics",
	month = jul,
	year = "2019",
	address = "Florence, Italy",
	publisher = "Association for Computational Linguistics",
	pages = "4762--4779",
}

@inproceedings{kemp,
	title={Knowledge Bridging for Empathetic Dialogue Generation},
	author={Li, Qintong and Zhang, Yizhe and Liang, Chenyang and Li, Nan and Li, Jianfeng},
	booktitle={Proceedings of the AAAI Conference on Artificial Intelligence},
	volume={35},
	number={18},
	pages={15727--15735},
	year={2021}
}

@inproceedings{speer2017conceptnet,
	title={Conceptnet 5.5: An open multilingual graph of general knowledge},
	author={Speer, Robyn and Chin, Joshua and Havasi, Catherine},
	booktitle={Proceedings of the AAAI conference on artificial intelligence},
	volume={31},
	number={1},
	year={2017}
}

@article{wang2025abkd,
  title={ABKD: Pursuing a Proper Allocation of the Probability Mass in Knowledge Distillation via $\alpha$-$\beta$-Divergence},
  author={Wang, Guanghui and Yang, Zhiyong and Wang, Zitai and Wang, Shi and Xu, Qianqian and Huang, Qingming},
  journal={arXiv preprint arXiv:2505.04560},
  year={2025}
}

@article{sanh2019distilbert,
	title={DistilBERT, a distilled version of BERT: smaller, faster, cheaper and lighter},
	author={Sanh, Victor and Debut, Lysandre and Chaumond, Julien and Wolf, Thomas},
	journal={arXiv preprint arXiv:1910.01108},
	year={2019}
}

@inproceedings{empsoa,
	title = "Don{'}t Lose Yourself! Empathetic Response Generation via Explicit Self-Other Awareness",
	author = "Zhao, Weixiang  and
	Zhao, Yanyan  and
	Lu, Xin  and
	Qin, Bing",
	booktitle = "Findings of the Association for Computational Linguistics: ACL 2023",
	month = jul,
	year = "2023",
	address = "Toronto, Canada",
	publisher = "Association for Computational Linguistics",
	url = "https://aclanthology.org/2023.findings-acl.843",
	doi = "10.18653/v1/2023.findings-acl.843",
	pages = "13331--13344",
}

@inproceedings{zhang2025towards,
  title={Towards multimodal empathetic response generation: A rich text-speech-vision avatar-based benchmark},
  author={Zhang, Han and Meng, Zixiang and Luo, Meng and Han, Hong and Liao, Lizi and Cambria, Erik and Fei, Hao},
  booktitle={Proceedings of the ACM on Web Conference 2025},
  pages={2872--2881},
  year={2025}
}


\end{document}